\documentclass[runningheads]{llncs}
\usepackage[T1]{fontenc}
\usepackage{graphicx}
\usepackage{times}
\usepackage{epsfig}
\usepackage{graphicx}
\usepackage{amsmath}
\usepackage{amssymb}
\usepackage{soul}
\usepackage{capt-of}
\usepackage{multirow}
\usepackage{adjustbox}
\usepackage[thinlines]{easytable}
\usepackage{gensymb}
\usepackage{booktabs}
\usepackage{subcaption}
\usepackage{mfirstuc}
\usepackage{array}
\usepackage{xcolor}

\newcommand{\ra}[1]{\renewcommand{\arraystretch}{#1}}
\newcolumntype{C}{>{\centering\arraybackslash}p{2cm}}

\usepackage{todonotes}

\makeatletter
\newcommand{\printfnsymbol}[1]{%
  \textsuperscript{\@fnsymbol{#1}}%
}
\makeatother

\newenvironment{icompact}{
  \begin{list}{$\bullet$}{
    \itemindent -.05em
    \parsep 0pt plus 1pt
    \partopsep 0pt plus 1pt
    \topsep 2pt plus 2pt minus 2pt
    \itemsep 0pt plus 1.3pt
    \parskip 0pt plus 2pt
    \leftmargin 0.13in}
       }
{\normalsize\end{list}}

\usepackage{xspace}
\newcommand\sys{\textsc{EdgeMixup}\xspace}

\begin{document}
\title{\sys: Improving Fairness  for  Skin Disease Classification and Segmentation}
\titlerunning{\sys}
%
\author{Anonymous Submission}
\author{
Haolin Yuan\inst{2}\printfnsymbol{1}\and 
Armin Hadzic\inst{1}\thanks{equal contribution} \and
William Paul\inst{1} \and
Daniella Villegas de Flores\inst{4} \and 
Philip Mathew\inst{1} \and
John Aucott\inst{4} \and
Yinzhi Cao\inst{2} \and
Philippe Burlina\inst{1,2,3}
}
\authorrunning{Haolin Yuan, Armin Hadzic, et. al.}
\institute{}

%
\institute{ Johns Hopkins University, Applied Physics Laboratory, Laurel, MD 
\and
 Johns Hopkins University, Dept. of Computer Science
 \and
 Johns Hopkins University, Malone Center for Engineering in Healthcare
 \and
 Johns Hopkins University, School of Medicine
}
\maketitle              
\vspace{-0.5cm}

\begin{abstract}
Skin lesions can be an early indicator of a wide range of infectious and other diseases. The use of deep learning (DL) models to diagnose skin lesions has great potential in assisting clinicians with prescreening patients. However, these models often learn biases inherent in  training data, which can lead to a performance gap in the diagnosis of people with light and/or dark skin tones. To the best of our knowledge, limited work has been done on identifying, let alone reducing, model bias in skin disease classification and segmentation. In this paper, we examine DL fairness and demonstrate the existence of bias in classification and segmentation models for subpopulations with darker skin tones compared to individuals with lighter skin tones, for specific diseases including Lyme, Tinea Corporis and Herpes Zoster. Then, we propose a novel preprocessing, data alteration method, called \sys, to improve model fairness with a linear combination of an input skin lesion image and a corresponding a predicted edge detection mask combined with color saturation alteration. For the task of skin disease classification, \sys outperforms much more complex competing methods such as adversarial approaches, achieving a 10.99\% reduction in accuracy gap between light and dark skin tone samples, and resulting in 8.4\% improved performance for an underrepresented subpopulation. 

\end{abstract}
\section{Introduction}
\label{sec:introduction}

Early detection of skin lesions can aid in identifying a range of infectious diseases. We consider Lyme disease~\cite{LymeHinckley,kuehn2013cdc}--- which affects nearly 476,000 cases per annum during 2010-2018~\cite{kugeler2021estimating}. Lyme disease is caused by the bacterium \textit{Borrelia burgdorferi}, which manifests via a red concentric lesion, called Erythema Migrans (EM), at the site of a tick bite~\cite{NADELMAN2015211}. While the EM pattern may appear simple to recognize, its diagnosis can be challenging for those with or without a medical background alike, as only 20\% of United States patients have the stereotypical bull's eye lesion~\cite{emTibbles}. When skin lesions are atypical they can be mistaken for other diseases such as Tinea Corporis (TC) or Herpes Zoster (HZ)~\cite{mazori2015vesicular}, two other diseases acting a confusers for Lyme,  considered herein. This has increased interest in medical applications of deep learning (DL), and using deep convolutional neural networks (CNNs), to assist clinicians in timely and accurate diagnosis of conditions including Lyme disease, TC and HZ~\cite{fujisawa2019deep,gu2019progressive,burlina2020ai}.
   
A major challenge in diagnosing skin diseases with CNNs is that they have been shown to learn and exhibit bias inherent in training data~\cite{hermann2020origins}. For example, the diagnostics accuracy of people with light skin is often higher than those with dark skin because a) the training may not have sufficient samples of dark skin with the condition, or b) there may exists an inherent correlation between image markers of protected factors and disease. In response, the AI community has been investigating bias mitigation strategies such as data generation for underrepresented subpopulations~\cite{paul2020TARA} or adversarial debiasing~\cite{zhang2018mitigating}. However, while applying CNNs to dermatology is of growing interest, insufficient attention has been directed towards identifying or reducing the prevalence of bias in CNN prediction for skin disease classification and segmentation. Existing bias mitigation strategies often perform poorly on skin diseases, especially for segmenting and classifying Erythema Migrans (EM), because they tend to remove important information on the lesion area or important image markers after debiasing. 
   
We propose a novel data preprocessing and alteration method, called \sys, to improve fairness in skin disease classification and segmentation. The key insight of this approach is to alter a skin image with a linear combination of the source image and a detected edge mask so that the lesion structure is preserved while minimizing skin tone information, which is done by altering the color composition in HSV space, thereby minimizing the ability of the model to infer information about the protected factor. This combined preprocessing approach, while simple, is shown to be significantly more effective than competing methods such as adversarial approaches which are also aimed at masking markers or protected factors. 
   
We evaluate \sys with fairness metrics for skin disease segmentation and classification tasks. First, for the segmentation task, we construct a dataset composed of 185 publicly available diseased skin images with  annotations for three regions: background, skin and lesion, conducted under clinician supervision and Institutional Review Boards (IRB) approval. Next, we demonstrate the existence of segmentation model bias on our annotated dataset. Our results show that \sys is able to reduce  bias to improve fairness and increase utility (as measured via $\mathtt{Jaccard}$ and $\mathtt{Dice}$). Second, for the classification task, we collect and have a clinician supervise annotation for a skin disease dataset with 2,712 (publicly-available) skin images classified into four classes, i.e., No Disease (NO), TC, HZ, and EM. We perform evaluation on the classification task using a traditional ResNet34 baseline and demonstrate the existence of significant bias. We show that \sys substantially improves model fairness compared to the baseline and also significantly outperforms state-of-the-art (SOTA) debiasing methods in improving performance on joint fairness-utility metrics.
%
%
Our contributions are:
\begin{icompact}
 \item We collect, annotate, and present two novel skin disease datasets with emphasis on Lyme disease, Tinea Corporis, and Herpes Zoster, for studying segmentation, classification, and addressing fairness, which we will publicly release upon publication. 
 \item We demonstrate for the {\it first} time that a segmentation model may exhibit bias for these important diseases.
 \item We propose \sys, a novel data preprocessing method that jointly addresses utility and fairness for the tasks of classification and segmentation  of skin diseases. 
 \item We evaluate \sys on skin lesion classification and segmentation, showing that it improves utility and fairness for segmentation and their tradeoff for classification, which also outperforms the SOTA approach. 
\end{icompact}

\section{Related Work}
\label{sec:related_work}

We provide an overview of prior work in skin disease classification and segmentation, as well as bias mitigation methods in the domain of medical imaging.

{\bf Skin Disease Classification and Segmentation:}
Deep CNNs have gained popularity for automated melanoma skin lesion segmentation due to disease relevance and model performance, despite the prevalence of fuzzy borders, inconsistent lighting conditions, and image artifacts~\cite{bi2017dermoscopic,zhang2019automatic}. Individual Topology Angle (ITA) has been used as a proxy for skin tone labels in medical imagery for segmentation and classification tasks. Little bias was found in skin disease segmentation and classification models using the SD-136~\cite{sun2016benchmark} and ISIC2018~\cite{codella2019skin} datasets~\cite{kinyanjui2020fairness}, which differ from the diseases this study focuses on. In this work, we reach the opposite conclusion for segmentation and classification of specific skin diseases and their lesions, including Lyme, EM, TC, and HZ. This also motivates the design of \sys in improving the fairness of skin lesion segmentation and classification. 


{\bf Bias Mitigation:}
Addressing bias in deep learning models can be categorized into three categories~\cite{caton2020fairness}: (1) {\it preprocessing}, such as augmentation and re-weighting; (2) {\it in-processing}, like adversarial debiasing; and (3) {\it post-processing}, such as thresholding. First, masking sensitive factors in imagery is shown to improve fairness in object detection and action recognition~\cite{Wang_2019_ICCV}. Second, adversarial debiasing operates on the principle of simultaneously training two networks with different objectives~\cite{goodfellow2014generative,madry2017towards,shafahi2019adversarial}. The competing two-player optimization paradigm is applied to maximizing equality of opportunity in~\cite{beutel2017data}.
This  technique has shown success for tabular data~\cite{zhang2018mitigating}, word embeddings~\cite{bolukbasi2016man}, and imagery~\cite{zhang2020towards}. Lastly,  Hardt el al.~\cite{hardt2016equality} adjust model outputs using thresholds to mitigate discrimination against a specified sensitive attribute.

By contrast, we propose \sys a much less complex but also more effective preprocessing approach to debiasing when applied to skin disease, and particularly Lyme-focused, classification and segmentation tasks.



\begin{figure}[!t]
\centering 
    \begin{subfigure}{0.38\textwidth}
    \centering
    \includegraphics[trim=0 230 0 0, clip, width=\linewidth]{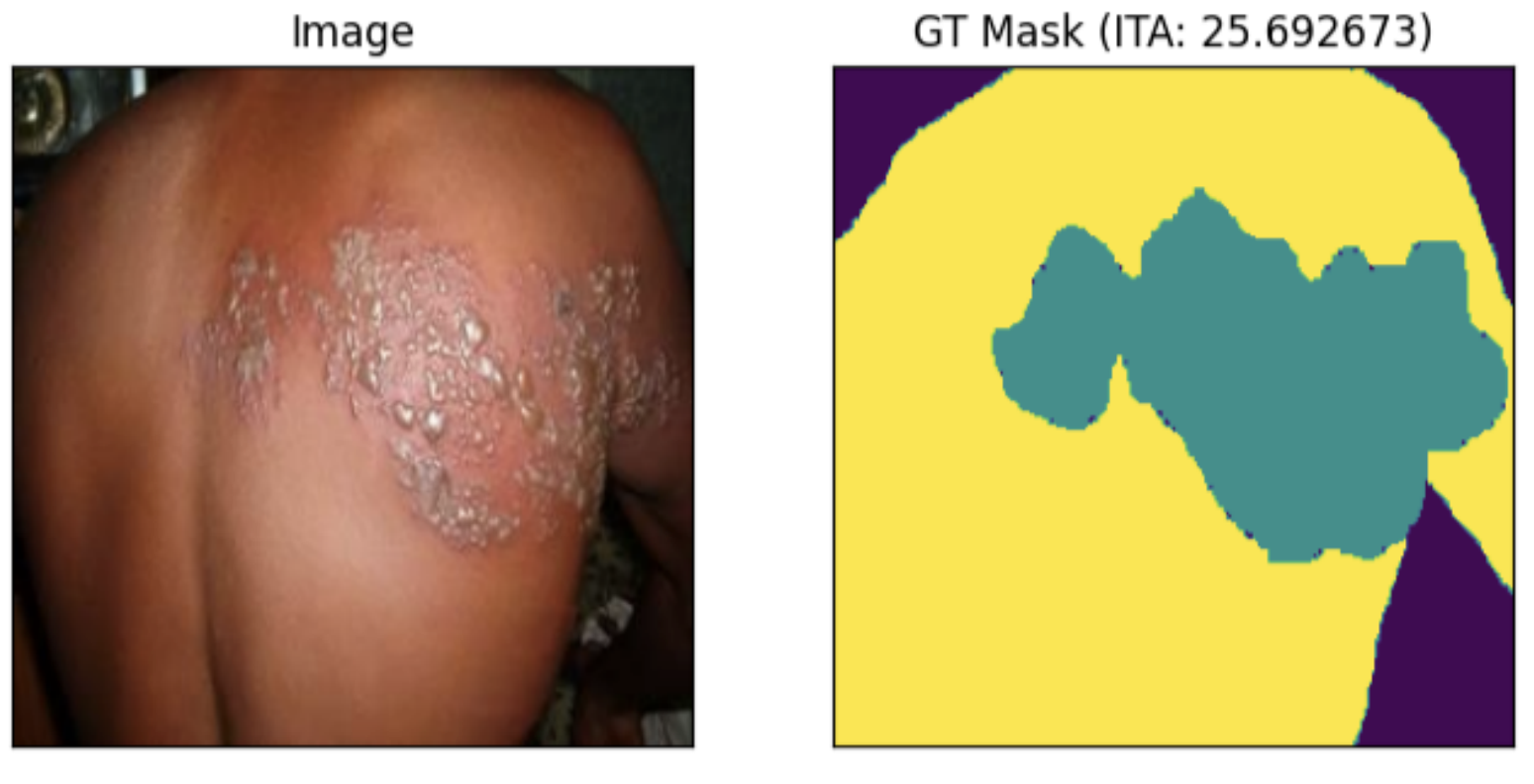}
    \end{subfigure}
    \hspace{0.03\textwidth}
    \vspace{0.03\textwidth}
    \begin{subfigure}{0.38\textwidth}
    \centering
    \includegraphics[trim=0 230 0 0, clip, width=\linewidth]{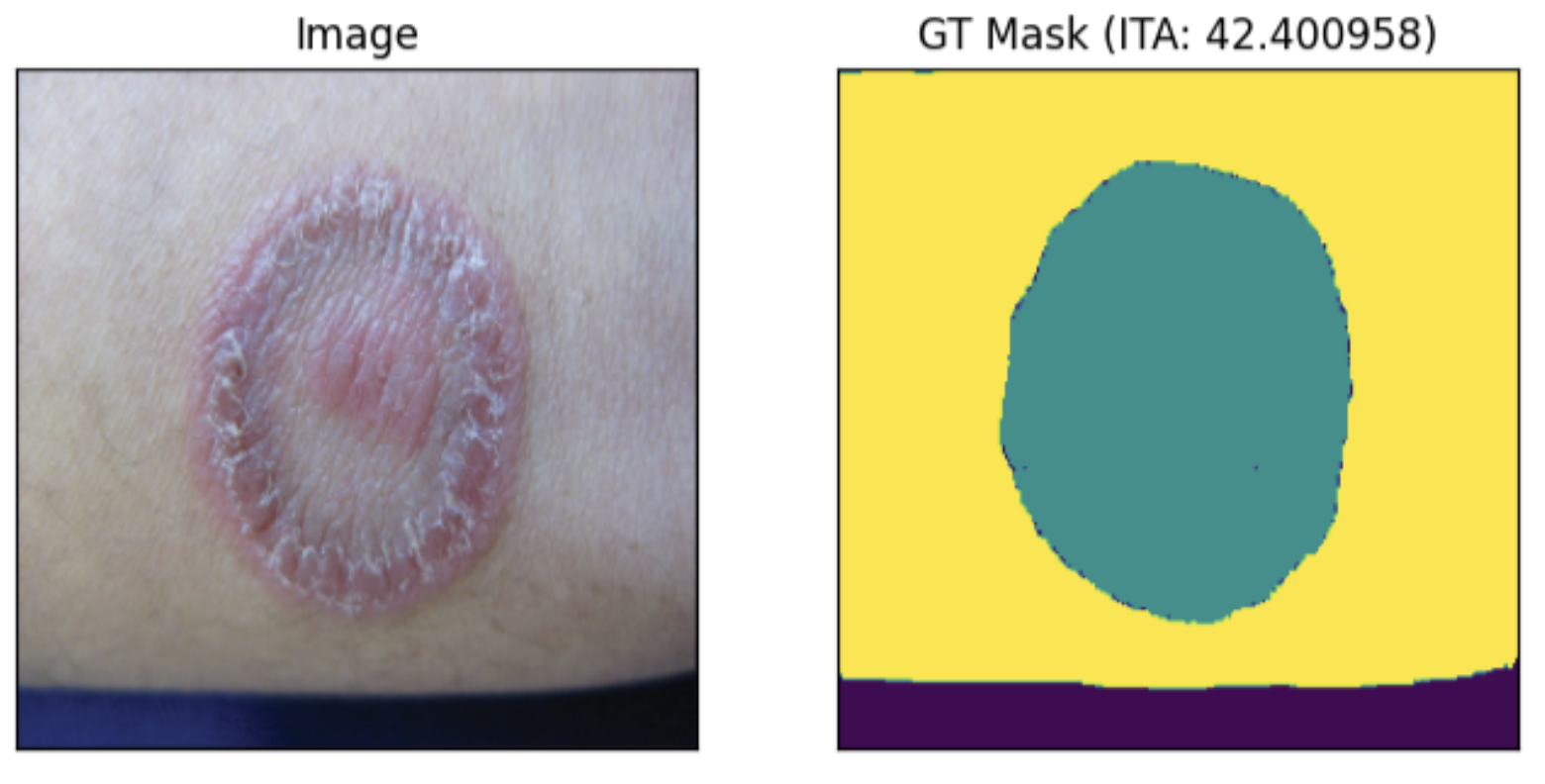}
    \end{subfigure}
    
    \vspace{-0.2cm}
    
    
    \vspace{-0.2in}
    \caption{Segmentation examples and ground truth (GT) manual annotations. Segmentation annotations (masks) show background (purple), skin (yellow), and lesions (blue).} \vspace{-0.2in}
    \label{fig:segmask_examples}
\end{figure}

\section{Datasets}
\label{sec:datasets}

We collect, annotate, and then present two datasets for skin disease segmentation and classification which we will publicly release upon publication.  
First, we collect skin images either from publicly available sources or from clinicians with patient informed consent.  Second, a medical technician and a clinician in our team manually annotate each image.  Data annotation follow the specific task/dataset as indicated below:

\begin{icompact}
\item {\bf Segmentation:}  We annotate skin images into three classes: background (black), skin (yellow), and lesion (blue), see Fig.~\ref{fig:segmask_examples}. The lesion area contains three types of disease/lesions: Tinea Corporis
(TC), Herpes Zoster (HZ), and Erythema Migrans
(EM). 
\item {\bf Classification:} We annotate skin images by classifying them into four classes: No Disease (NO), TC, HZ, and EM. 
\end{icompact}

Table~\ref{tbl:segsplits} shows the characteristics of these two datasets broken down by the disease type and skin tone, as calculated by the Individual Typology Angle (ITA)~\cite{wilkes2015fitzpatrick}. Specifically, we consider tan2, tan1, and dark as dark skin (ds) and others as light skin (ls). One prominent observation is that ls images are more abundant than ds images due to a disparity in the availability of ds imagery found from either public sources or from clinicians with patient consent. This disparity motivates the design of \sys in improving model fairness in diagnosing skin diseases.

\begin{table}[!t]
  \renewcommand{\arraystretch}{0.8} 
  \setlength{\tabcolsep}{2.8pt}
    \caption{Annotated segmentation and classification dataset Characteristics, broken down by ITA-based skin tones (ls: light skin and ds: dark skin) and disease types (EM:  Erythema Migrans, HZ:  Herpes Zoster, TC: Tinea Corporis, and NO: No disease).}  	 \label{tbl:segsplits}
	\centering
	\begin{tabular}{cccccccccccccccccccc}
	\toprule
	\multirow{4}{*}{Split} & \multicolumn{9}{c}{Segmentation Dataset} & \multicolumn{10}{c}{Classification dataset} \cr
	  \cmidrule(lr){2-10} \cmidrule(lr){11-20}
     & NO & \multicolumn{2}{c}{EM} & \multicolumn{2}{c}{HZ} &\multicolumn{2}{c}{TC} & \multicolumn{2}{c}{Total} & \multicolumn{2}{c}{NO} & \multicolumn{2}{c}{EM} & \multicolumn{2}{c}{HZ} & \multicolumn{2}{c}{TC} & \multicolumn{2}{c}{Total} \cr
     \cmidrule(lr){2-2} \cmidrule(lr){3-4} \cmidrule(lr){5-6} \cmidrule(lr){7-8} \cmidrule(lr){9-10} \cmidrule(lr){11-12} \cmidrule(lr){13-14} \cmidrule(lr){15-16} \cmidrule(lr){17-18} \cmidrule(lr){19-20} 
    & ls/ds & ls & ds & ls & ds & ls & ds & ls & ds & ls & ds & ls & ds & ls & ds & ls & ds & ls & ds \cr
    \midrule
    Train & 0 & 35 & 3 & 34 & 4 & 28 & 9 & 97 & 16 & 700 & 57 & 578 & 41 & 522 & 75 & 517 & 84 & 2,317 & 257 \cr
    Val   & 0 & 5  & 5 & 5  & 5 & 5  & 5 & 15 & 15 & 36  & 2  & 31  & 10 &  35 & 6  & 26  & 5  & 128  & 23 \cr
    Test  & 0 & 7  & 7 & 7  & 7 & 7  & 7 & 21 & 21 & 86  & 4  & 73  &  7 &  51 & 9  & 66  & 6  & 276  & 26 \cr
	\bottomrule
	\end{tabular} 

\end{table} 


\section{Method}
\label{sec:methods}

We present our core method in reducing skin tone bias for segmentation and classification CNNs.  We start by describing the design of \sys, and then present how we apply \sys for the tasks of segmentation and classification.



\subsection{EdgeMixup data preprocessing}

\begin{figure}[tb]\captionsetup{font=footnotesize}
    \centering 
    \begin{subfigure}{0.19\textwidth}
        \includegraphics[width=\linewidth]{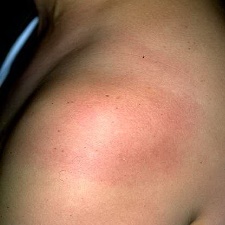}
        \caption{Original Image}\label{fig:diff_operator-1}
    \end{subfigure}
    \begin{subfigure}{0.19\textwidth}
        \includegraphics[width=\linewidth]{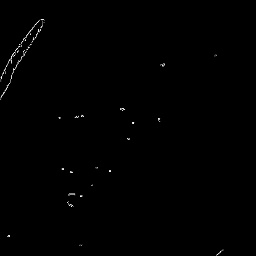}
        \caption{Canny}\label{fig:diff_operator-2}
    \end{subfigure}
    \begin{subfigure}{0.19\textwidth}
        \includegraphics[width=\linewidth]{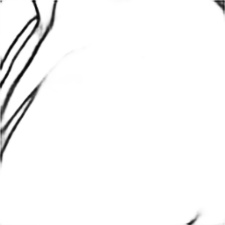}
         \caption{DexiNed-fused} \label{fig:diff_operator-3}
    \end{subfigure}
    \begin{subfigure}{0.19\textwidth}
        \includegraphics[width=\linewidth]{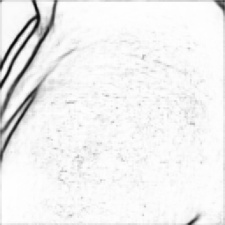}
        \caption{DexiNed-avg} \label{fig:diff_operator-4}
    \end{subfigure}
    \begin{subfigure}{0.19\textwidth}
        \includegraphics[width=\linewidth]{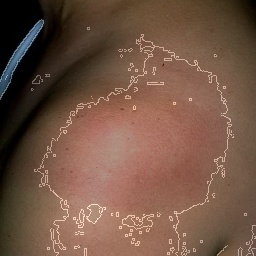}
         \caption{EdgeMixup} \label{fig:diff_operator-5}
    \end{subfigure}
    
    \caption{A motivating example to compare different edge detection methods}
    \label{fig:diff_operator}\vspace{-0.1in}
\end{figure}

The key insight of \sys is to ``mix-up'' a detected edge image with the original skin image for data preprocessing via a linear combination.  Intuitively, such preprocessing not only highlights the skin lesion, via an edge image, but also suppresses the skin tone. While this idea is intuitively simple, the edge detection is challenging due to color similarity causing ambiguous edges between skin and lesions.  
We start with a motivating example to illustrate this challenge. 

{\bf Motivating example:} Fig.~\ref{fig:diff_operator-1} shows a skin disease image with atypical EM, in which the lesion has no clear boundary with the skin, making the edge detection challenging.  We test two different edge detectors on this scenario: the Canny operator~\cite{canny_operator} and the SOTA DL-based edge detector DexiNed~\cite{dexined} (trained on BIPED~\cite{BIPED}). DexiNed includes two version: fused (
concatenation and fusion of all predictions from the neural network) and avg (an average of all predictions). Fig.~\ref{fig:diff_operator-2}--\ref{fig:diff_operator-4} shows the edge detection result. Clearly, Canny fails to even detect a basic human silhouette; DexiNed-fused detects some of the human body's edge, but not the lesion's. DexiNet-avg is better at detecting some parts of the lesion, but not its edge. As a comparison, we also depict the edge detection of \sys in Fig.~\ref{fig:diff_operator-5}, which clearly shows the lesion boundary.


\begin{figure}[!t]
        \centering
        \includegraphics[width=\textwidth]{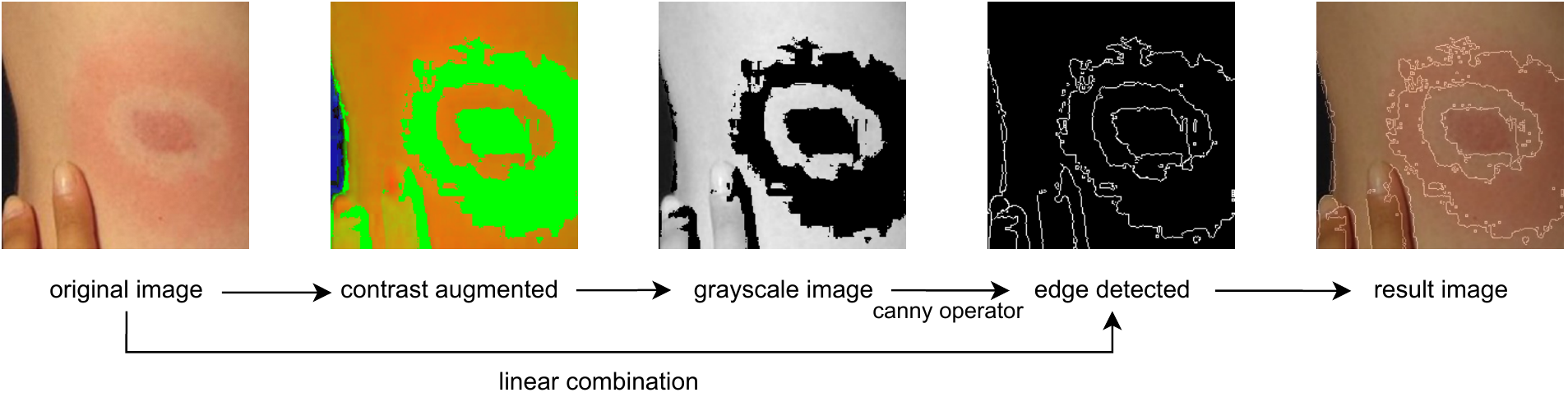}
 \caption{EdgeMixup Process}   
 \label{fig:fig_process}
 \vspace{-0.2in}
\end{figure}


{\bf Approach:}
Fig.~\ref{fig:fig_process} summarizes the overall process of \sys's data preprocessing into four steps.  First, \sys converts a given image to the Hue-Saturation-Value (HSV) color space. Then, \sys applies a red mask in the HSV color space to zero-out the red and blue channels and maximize the green color to 255. The image from this step is called contrast augmented. Second, \sys selects the value (V), or lightness, channel of the contrast augmented image from the previous step to produce a gray-scale image. Third, \sys applies a Canny edge detector to extract edge boundary and generates an edge image.  Lastly, \sys combines the edge image and the original sample image linearly, like a mixup, to generate an altered image (called a result image). If not otherwise specified, the default weight for edge image in the linear combination is 0.3.

\subsection{Application of \sys on different diagnostics-related tasks:}

The purpose of \sys is to improve fairness in diagnostics models via data alteration and pre-processing. Next, we apply \sys to two types of DL-based diagnostics tasks with the aim of improving fairness.

{\bf Lesion Segmentation:} Lesion segmentation aims to separate a skin lesion from regular skin to assist clinicians in the examination and diagnosis of EM by simplifying time-series clinical comparisons. \sys preprocesses training images before feeding them into a segmentation model (both at training and inference time), e.g. U-Net~\cite{ronneberger2015u}, which then segments the images into three regions: background, skin, and lesion.

{\bf Disease Classification:} Disease classification aims to prescreen and diagnose, principally, EM (for Lyme Disease), and also classify possible Lyme confusers including: Tinea Corporis
(TC), Herpes Zoster (HZ), and no disease (NO). Again, \sys alters the original training images prior to training a classification model, such as a ResNet34~\cite{resnet}.

\section{Experimental Setup}


\subsection{Lesion Segmentation}

Our evaluation baseline is a U-Net trained to segment images of skin lesions into three categories: background, skin, and lesion. Our evaluation metrics include metrics for utility and fairness, since often (but not always) these two may tradeoff. Utility is measured using a $\mathtt{Jaccard}$ score~\cite{Jaccard} and $\mathtt{Dice}$ coefficient~\cite{Dice}, which measure the similarity between a predicted mask and the manually annotated ground truth. Higher similarity results in higher the model performance. Fairness is evaluated by the gap of the $\mathtt{Jaccard}$ score and $\mathtt{Dice}$ coefficient between ls and ds images, notated as $\mathtt{J}_\mathtt{gap}$ and $\mathtt{D}_\mathtt{gap}$ respectively. The smaller the gap is, the more fair the model. 



\subsection{Disease Classification}

\subsubsection{Baselines:} We select ResNet34 as a baseline model, with ImageNet pretrained weights, early stopping, and a learning rate of 1e-3 trained for 100 epochs. Our evaluation for classification debiasing involves the following competing debiasing approaches:

\begin{icompact}
\item Adversarial Debiasing (AD). \hspace{0.05in} An in-processing method~\cite{zhang2018mitigating} using adversarial debiasing, where a separate classifier/player is tasked to predict the protected factor using the true class and the task prediction classifier's internal representation of a given input image.
\item Mask. \hspace{0.05in} A mask-based debiasing approach leveraging a synthesized mask from a segmentation network to mask out skintone in images input to the classifier. 
\item Mask+AD. \hspace{0.05in} A combination of Mask and AD aimed at masking skintone information both in image and embedded space.
\item DexiNed-avg. \hspace{0.05in} DexiNed-avg entails the use of the average version of DexiNed~\cite{dexined} as an edge detector used by \sys. 
\end{icompact}


\subsubsection{Evaluation Metrics:} We use the following disease classification evaluation metrics. 
\begin{icompact}
 \item Accuracy based metrics: We measure  accuracy to characterize utility. To measure fairness we use accuracy gap between ls and ds subpopulations, and the (Rawlsian) minimum accuracy across subpopulations. To characterize the  tradeoff between utility and and fairness we use the joint metric from~\cite{paul2020TARA}:
 \begin{equation}
\mathtt{CAI}_{\alpha} = \alpha (\mathtt{acc}_\mathtt{gap}^\mathtt{baseline} - \mathtt{acc}_\mathtt{gap}^\mathtt{debiased}) + (1 - \alpha) (\mathtt{acc}^\mathtt{debiased} - \mathtt{acc}^\mathtt{baseline}).
\end{equation}
\item AUC (Area under the receiver operating characteristic curve): Similarly, we also measure utility with AUC, and fairness via AUC gap and minimum AUC. Likewise, following prior work~\cite{paul2020TARA} we also calculate the AUC-based joint utility/fairness metric defined as:
\begin{equation}
\mathtt{CAUCI}_{\alpha} = \alpha (\mathtt{AUC}_\mathtt{gap}^\mathtt{baseline} - \mathtt{AUC}_\mathtt{gap}^\mathtt{debiased}) + (1 - \alpha) (\mathtt{AUC}^\mathtt{debiased} - \mathtt{AUC}^\mathtt{baseline}).
\end{equation}
\end{icompact}

\section{Results}
\label{sec:discussion}

In this section, we present results on on the task of lesion segmentation and skin disease classification. We also evaluate the performance and fairness of \sys compared with 
%
%
adversarial debiasing (AD), synthesized masking (Mask) in terms of fairness improvement and both (Mask+AD). Note, our code will be released upon publication. 


{\bf Skin Lesion Segmentation:}
Table~\ref{tbl:SegBias} shows the performance of \sys and a baseline U-Net on our segmentation dataset. We compare predicted masks with the manually-annotated ground truth by calculating the $\mathtt{Jaccard}$ and $\mathtt{Dice}$ scores, and computing the gap for each of the two scores for subpopulations with ls and ds (based on ITA). The results present two clear findings. First, \sys, as a data preprocessing method, improves the utility of lesion segmentation in terms of $\mathtt{Jaccard}$ and $\mathtt{Dice}$. A possible reason is that \sys clearly preserves key skin lesion information, thus improving the segmentation quality, while attenuating markers for protected factors.  Likewise, \sys also improves the fairness of the segmentation task by lowering the gap of the $\mathtt{Jaccard}$ and $\mathtt{Dice}$ scores  between people with ls and ds. As a result, \sys demonstrates consistency in improving both utility as well as fairness in term of the utilized metrics. 



\begin{table*}[!t]
\caption{Segmentation Test Results: Performance and Fairness}
\scriptsize
    \begin{center}
        \ra{1.3}
        \begin{tabular}{@{} l C C C C C @{}}
        \toprule
             Method & $\mathtt{Jaccard}$ & $\mathtt{J}_\mathtt{gap}$ & $\mathtt{Dice}$  & $\mathtt{D}_\mathtt{gap}$\\ 
        \midrule 
            U-Net (Baseline) & 0.7053 (0.0035) & 0.0809 (0.0001) & 0.8225 (0.0029)  & 0.0743 (0.0002) \\ 
            {\bf \sys} & \textbf{0.7597 (0.0033)} & \textbf{0.0422 (0.0001)} & \textbf{0.8609 (0.0026)}  & \textbf{0.0295 (0.0001)} \\ 
        \bottomrule
        \end{tabular}
    \end{center}
\label{tbl:SegBias} \vspace{-0.2in}
\end{table*}

{\bf Skin Disease Classification:}
Table~\ref{tbl:ClassificationResults} shows utility performance ( $\mathtt{acc}$ and $\mathtt{AUC}$) and fairness results (gaps of $\mathtt{acc}$ and $\mathtt{AUC}$ between ls and ds subpopulations). Note that we list the margin of error of each number in the parenthesis. Clearly, \sys outperforms SOTA approaches in balancing the model's performance and fairness, i.e., the $\mathtt{CAI}_{\alpha}$ and $\mathtt{CAUCI}_{\alpha}$ values of \sys are the highest compared with the vanilla ResNet34 and other baselines. Next, we examine the different metrics separately. 

First, the $\mathtt{acc}$ value of \sys is the second largest, which is only second to the baseline ResNet34, but higher than all other competing debiasing methods. While a decrease in utility often arises for debiasing, our results show that \sys is effective in largely preserving the model's utility. The $\mathtt{acc}_\mathtt{gap}$ is also the second smallest, which is only second to the Mask approach (i.e., applying a segmentation mask to disease images). Note that the $\mathtt{acc}$ value of the Mask approach is a mere 73.84\%, suggesting utility was substantially sacrificed to maximize fairness. Next, the $\mathtt{acc}_\mathtt{min, ds}$  of \sys is the highest among all approaches, meaning that \sys is superior at improving classification performance for underrepresented ds subgroups. 

Second, the $\mathtt{AUC}$ value of \sys is only around 1\% smaller than of the baseline ResNet34 model, highlighting \sys's strong performance in disease classification. At the same time, the  $\mathtt{AUC}_\mathtt{gap}$ is the smallest among all approaches, while $\mathtt{AUC}_\mathtt{min, ds}$ is the largest. This showcases that \sys has the best characteristics in terms of fairness, as well as addressing the overall fairness/utility criterion; thus, improving the overall system performance.


\begin{table*}[!t]
\scriptsize
    \caption{Skin disease classification and associated bias. Sample types include normal skin (NO) and diseased skin: EM, HZ, and TC. Samples contain skin tones as a protected factor.}\vspace{-0.1in} \label{tbl:ClassificationResults}
    \begin{center}
        \ra{1.3}
        \begin{tabular}{@{}lcccccc@{}}
        \toprule
            \multirow{2}{*}{Metrics}& \multirow{2}{*}{ResNet34} & \multicolumn{4}{c}{Baselines} & \multirow{2}{*}{\bf EdgeMixup} \\
            \cmidrule(lr){3-6}
             &  & AD & Mask & Mask+AD &  DexiNed-avg &    \\
        \midrule
            $\mathtt{acc}$ & \textbf{85.10 (4.02)} & 81.79 (4.35) & 73.84 (4.96) & 72.52 (5.04) & 69.87 (5.17) & 83.44 (4.19)\\
            $\mathtt{acc}_\mathtt{gap}$ & 13.15 (12.98) & 5.33 (11.69) & \textbf{0.83 (11.87)} & 4.82 (10.91) & 19.79 (13.52) &2.16 (10.28) \\
            $\mathtt{acc}_\mathtt{min, ds}$ & 73.08  & 76.92  & 73.08  & 72.10  & 51.85 & \textbf{81.48} \\
            $\mathtt{CAI}_{0.5}$ & - & 2.2550 & 0.5300 & -2.1250 & -10.9350 &\textbf{4.6650} \\
            $\mathtt{CAI}_{0.75}$ & - & 5.0375 & 6.4250 & 3.1025 & -8.7875 &\textbf{7.8275} \\
            \midrule
            $\mathtt{AUC}$ & \textbf{0.9725 (0.0185)} & 0.9555 (0.0233) & 0.9072 (0.0327) & 0.9053 (0.0330) & 0.8892 (0.0354)  &0.9623 (0.0215)\\
            $\mathtt{AUC}_\mathtt{gap}$ & 0.0331 (0.0714) & 0.0094 (0.0469) & 0.0275 (0.0618) & 0.0414 (0.0536) & 0.0898 (0.1129) &\textbf{0.0076 (0.0554) } \\
            $\mathtt{AUC}_\mathtt{min, ds}$  & 0.9420  & 0.9548 & 0.9050 & 0.9023 &  0.8069  &\textbf{0.9556 } \\
            $\mathtt{CAUCI}_{0.5}$ & - & 0.0034 & -0.0299 & -0.0378 & -0.070000&\textbf{0.007650} \\
            $\mathtt{CAUCI}_{0.75}$ & - & 0.0135 & -0.0121 & -0.0230 & -0.063350 &\textbf{0.016575} \\
        \bottomrule
        \end{tabular}
    \end{center} \vspace{-0.3in}
\end{table*}

\section{Discussion}

Our study, performed under IRB approval (and to be publicly released), demonstrates for the first time the possible presence of bias when addressing Lyme disease, and other important conditions that act as confusers to Lyme (HZ and TC) when using a vanilla classifier. A fact, never reported before and also in contrast to other skin diagnostic studies. This observation highlights the importance of studying skin disease bias with datasets that have much larger exemplar cardinality for Lyme, HZ, and TC when compared to the other prevalent datasets, such as SD-198, that may not focus as much on those diseases. We also present a simple, yet highly effective, method to debias models, and show how the method produces a censoring/masking effect, vis-a-vis protected attributes markers, without a debilitating effect on utility.

\section{Conclusion}
\label{sec:conclusion}

We present a study to identify, quantify, and mitigate bias of skin image classification and segmentation models trained from two datasets collected in our study. 
Specifically, we propose \sys, a novel data preprocessing method that utilizes edge detection to isolate skin lesions.
 \sys outperforms the previous SOTA (81.58\%) by 1.86\% accuracy and other debiasing methods with a $\mathtt{CAI}_{0.5}$ of 4.6650. We adapt \sys for the task of skin lesion segmentation on our new dataset and surpass the baseline method by 0.0544 in $\mathtt{Jaccard}$ score and reduce the $\mathtt{Jaccard}$ gap ($\mathtt{J}_\mathtt{gap}$) in performance between the light and dark skin subpopulations by 0.0387. \sys is an effective approach that achieves fair performance across subpopulations with respect to skintone.



\newpage

{\small
\bibliographystyle{splncs04}
\bibliography{egbib}
}

\newpage

%
%
%
%

\end{document}